%% file: ms.tex
\documentclass[letterpaper, 10 pt, conference]{ieeeconf}  

\IEEEoverridecommandlockouts                              

\usepackage{xcolor}
\usepackage{soul}
\usepackage[utf8]{inputenc}
\usepackage[small]{caption}
\usepackage{booktabs}
\usepackage{graphics} 
\usepackage{tabularx}
\usepackage{amsmath} 
\usepackage{amssymb}  
\usepackage{mathrsfs}
\usepackage{subcaption}
\usepackage{tikz}
\usetikzlibrary{shapes,arrows}
\usepackage{pgfplots}
\usepackage[per-mode=symbol]{siunitx}
\usepackage{url}
\usepackage[noadjust]{cite}
\usepackage{bm}
\newcommand{\vect}[1]{\boldsymbol{\mathbf{#1}}}
\renewcommand{\arraystretch}{1.2}
\newcommand{\ra}[1]{\renewcommand{\arraystretch}{#1}}
\pgfplotsset{compat=newest}
\pgfplotsset{every axis/.append style={
	font=\LARGE}
}
\pgfplotsset{every axis legend/.append style={legend cell align=left}}

\def\man#1;{%
    \begin{scope}[shift={#1}]
        \fill [rounded corners=1.5] (0,0.4) -- (0,0.8) -- (0.4,0.8) -- (0.4,0.4) --
            (0.325,0.4) -- (0.325,0.7) -- (0.3,0.7) -- (0.3,0) -- (0.225,0) --
            (0.225,0.4) -- (0.175,0.4) -- (0.175,0) -- (0.1,0) -- (0.1,0.7) --
            (0.075,0.7) -- (0.075,0.4) -- cycle;
        \fill (0.2,0.9) circle (0.1);
    \end{scope}}

\usetikzlibrary{arrows.meta, calc, shapes}
\tikzset{%
  >={Latex[width=2mm,length=2mm]},
            base/.style = {rectangle, rounded corners, draw=black,
                           minimum width=1cm, minimum height=1cm,
                           text centered, font=\sffamily},
            simulator/.style = {base, fill=green!30, minimum width=4cm},
            solver/.style = {base, fill=red!30},
            reward/.style = {base, minimum height=1.5cm},
            module/.style = {base, minimum width=2.5cm, minimum height=1.5cm, fill=blue!30},
            module2/.style = {base, minimum width=2.5cm, minimum height=1.5cm, fill=white},
}
\pgfdeclarelayer{background}
\pgfdeclarelayer{foreground}
\pgfsetlayers{background,main,foreground}

\def\ysplit{1.0}
\def\xsplit{2.5}

\makeatletter
\let\NAT@parse\undefined
\makeatother
\usepackage{hyperref} 
\usepackage{cleveref}

\usepackage[keeplastbox]{flushend}

\title{\Large \bf Adaptive Stress Testing for Autonomous Vehicles}

\author{Mark Koren,$^{1*}$ Saud Alsaif,$^{2*}$ Ritchie Lee,$^{3}$ and Mykel J. Kochenderfer$^{1}$
\thanks{*These authors contributed equally.}
\thanks{$^{1}$Mark Koren and Mykel J. Kochenderfer are with Aeronautics and Astronautics, Stanford University, Stanford, CA 94305, USA
        {\tt\small \{mkoren, mykel\}@stanford.edu}}
\thanks{$^{2}$Saud Alsaif is with Electrical Engineering, Stanford University, Stanford, CA 94305, USA
        {\tt\small salsaif@stanford.edu}}%
\thanks{$^{3}$Ritchie Lee is with Electrical and Computer Engineering, Carnegie Mellon University Silicon Valley, Moffett Field, CA 94035, USA 
        {\tt\small rititchie.lee@sv.cmu.edu}}
}

\begin{document}

\maketitle
\thispagestyle{empty}
\pagestyle{empty}

\begin{abstract}  
This paper presents a method for testing the decision making systems of autonomous vehicles. Our approach involves perturbing stochastic elements in the vehicle's environment until the vehicle is involved in a collision. Instead of applying direct Monte Carlo sampling to find collision scenarios, we formulate the problem as a Markov decision process and use reinforcement learning algorithms to find the most likely failure scenarios.  This paper presents Monte Carlo Tree Search (MCTS) and Deep Reinforcement Learning (DRL) solutions that can scale to large environments. We show that DRL can find more likely failure scenarios than MCTS with fewer calls to the simulator. A simulation scenario involving a vehicle approaching a crosswalk is used to validate the framework. Our proposed approach is very general and can be easily applied to other scenarios given the appropriate models of the vehicle and the environment.
\end{abstract}

\section{INTRODUCTION}\label{sec:intro}

While major advances have been made in improving the capabilities of decision making systems for automated vehicles, validation of these systems is challenging due to the vast space of driving scenarios~\cite{kalra2016driving, koopman2016challenges, wachenfeld2017new}.
Establishing confidence in any automotive system will involve road tests, but road tests alone do not adequately cover the space of critical scenarios~\cite{wachenfeld2017thesis, winner2015quo, zhao2017accelerated}.
Hence, research efforts have focused on validation by simulation with a particular emphasis on building realistic models of the environment, including models of driving behavior and sensors~\cite{woodrooffe2014performance, yang2010development, huang2017cutin, sunberg2017value, wheeler2017radar}.

In the validation of autonomous vehicles, it can be very valuable to know the most-likely failure scenarios as predicted by a probabilistic model of the environment. Direct sampling is inefficient due to the rarity of failure events. Adaptive stress testing (AST) has been proposed as a practical approach to finding most-likely failure scenarios by using a Markov decision process (MDP) formulation. The AST approach was first applied to test an aircraft collision avoidance system~\cite{lee2015adaptive}. The original method uses Monte Carlo tree search (MCTS) with double progressive widening (DPW)~\cite{couetoux2011continuous} to search for the most-likely failure condition. This paper adapts this approach to the automotive setting where the system under test (SUT) is an autonomous vehicle with noisy sensors approaching a pedestrian crosswalk. This paper also proposes deep reinforcement learning (DRL) as an alternative solver for AST.

This paper makes the following contributions:
\begin{itemize}
	\item We extend the AST methodology by introducing a new solver technique, DRL. We show that the theoretical advantages of AST still apply to the more general formulation.
	\item We present a simulation framework for autonomous vehicles that interfaces with our AST implementation. The framework is modular, which enables components such as decision making systems, sensor models, and simulation dynamic models to be interchanged with alternative and more sophisticated versions.
	\item We apply AST to a set of autonomous vehicle scenarios and show that AST can successfully find high probability failure scenarios.  Our experiments show that DRL can find better solutions much more efficiently compared to MCTS. 
\end{itemize}

The remainder of the paper is organized as follows. \Cref{sec:method} provides a review of MDPs, the two reinforcement learning algorithms used in this paper, and the AST framework. \Cref{sec:astav} introduces the specific formulation of AST for autonomous vehicles as well as the problem setup and evaluation metrics. \Cref{sec:res} presents the experimental results and analysis of performance. \Cref{sec:conc} concludes the paper.
\section{BACKGROUND AND METHODOLOGY}
\label{sec:method}
We present background material on the MDP formulation and the solvers we use along with the AST methodology. We also extend the AST methodology to include DRL.
\subsection{Markov Decision Processes}
\label{sec:MDPs}
In a Markov decision process (MDP), the agent chooses an action $a$ based on a state $s$  and receives a reward $r$ according to the reward function $R(s,a)$~\cite{DMU}. The state transitions to the next state $s'$ stochastically according to the state transition function $T(s' \mid s, a)$. The probability of transitioning to state $s'$ depends only on $s$ and $a$, which is known as the Markov assumption. The goal of an agent is to find a policy $\pi$ that specifies the action $a = \pi(s)$ at each state to maximize the expected utility. The utility of executing a policy $\pi$ from state $s$ is given recursively by the value function: 
\begin{equation}
V^{\pi} \left(s\right) = R\left(s, \pi\left(s\right)\right) + \gamma  \sum_{s'}T\left(s'\mid s,\pi \left(s\right)\right)V^{\pi}\left(s'\right)
\end{equation}
where $\gamma$ is the discount factor that controls the weight of future rewards.  Reinforcement learning algorithms, such as MCTS and DRL, can be used to find the optimal policy $\pi(s)$.

\subsection{Monte Carlo Tree Search}
MCTS is an online sampling-based algorithm that can be used to solve  MDPs~\cite{browne2012survey}. MCTS builds a search tree by sampling the state space and using forward simulation to estimate the value of states and actions. This paper uses a variation of MCTS with double progressive widening (DPW)~\cite{couetoux2011continuous}.   DPW regulates branching in the search tree to prevent the number of children from exploding when the number of states or actions is very large.

\subsection{Deep Reinforcement Learning}\label{sec:drl}
Deep reinforcement learning (DRL) is an alternative approach to solving MDPs that uses a feed-forward neural network to represent the policy $\pi_\theta(s)$. The policy is parameterized by $\theta$, which represents the neural network weights.  In our implementation, the policy maps an input state to the mean of a Gaussian distribution. The actions are then sampled from the distribution $a \sim \mathcal{N}(\pi_\theta(s), \Sigma)$, with the diagonal covariance matrix $\Sigma$ separately parametrized and independent of state \cite{schulman2015trust}.

To update the policy, we use Generalized Advantage Estimation (GAE) to estimate the policy-gradient from batches of simulation trajectories \cite{schulman2015high}. GAE defines an advantage function 
\begin{equation}
A^{\pi,\gamma} \left(s, a\right) := Q^{\pi,\gamma}\left(s, a\right)-V^{\pi,\gamma}\left(s\right)
\end{equation}
where $\gamma$ is the discount factor that governs the weight of future rewards.
The Q-function evaluates the value of taking an action from a state and then following the policy, and is defined as 
\begin{align}
Q^{\pi,\gamma}\left(s, a\right) &=  R\left(s,a\right)\notag \\
&+ \gamma  \sum_{s'}T\left(s'\mid s,a\right)Q^{\pi,\gamma}\left(s',  \pi\left(s'\right)\right)
\end{align}
Once the policy-gradient is known, Trust Region Policy Optimization (TRPO) is used to step the policy. TRPO generally gives monotonic increases in policy performance by constraining the KL divergence~\cite{schulman2015trust}.

\subsection{Adaptive Stress Testing}\label{sec:ast}
We formulate the problem of finding failure events as a sequential decision process, following prior work~\cite{lee2015adaptive}. The inputs to the problem are the pair ($\mathscr{S}$, $E$), where $\mathscr{S}$ is a generative simulator that is treated as a black box and $E$ is a subset of the state space where the event of interest (e.g. a collision) occurs. The simulator contains the models for the SUT, the models of the other agents in the environment, and the dynamics of the environment. The simulator exposes the following simulation control functions to the solver:
\begin{itemize}
\item \textsc{Initialize}$(\mathscr{S})$: Resets $\mathscr{S}$ to its initial state $s_0$.
\item \textsc{Step}$(\mathscr{S}, E, a)$: Steps the simulation in time by drawing the next state $s'$ after taking action $a$. The function returns the probability of the transition and an indicator whether $s'$ is in $E$ or not.
\item \textsc{IsTerminal}$(\mathscr{S}, E)$: Returns true if the current state of the simulation is in $E$, or if the horizon of the simulation $T$ has been reached. 
\end{itemize}
This formulation assumes that the state of the simulator is hidden, which in general means that the simulator is non-Markovian from the point of the view of the solver.  However, because we have chosen our actions $a$ to fix the stochastic elements of the simulator, the state transitions now become deterministic.  In this setting, we can deterministically revisit a state $s$ by replaying the history of actions that led to the state starting from the initial state. As such, we can use the sequence of actions $a_{0:t-1}$ to represent the state $s_t$~\cite{lee2015adaptive}. The abstraction allows us to overcome partial observability in the simulator. 

The definition of the problem is as follows: Given a simulator $\mathscr{S}$ and a subset of the state space $E$, find the most likely trajectory that leads to an event in $E$.  Because $a$ controls the stochastic elements of the simulator, the simulation does not evolve stochastically.  Rather, the actions controlling the adversarial environment uniquely determine the evolution of the scenario.

\begin{figure}[htbp]
	\centering
    \scalebox{0.8}{\input{ASTStruct.tex}}
    \caption{The AST methodology. The simulator is treated as a black box. The solver optimizes a reward based on transition likelihood and whether an event has occurred.}
	\label{fig:ASTStruct} 
\end{figure}
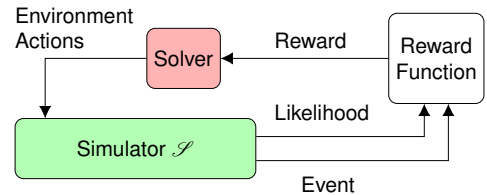

The approach to solve the AST problem is shown in~\Cref{fig:ASTStruct}. We start with the solver, which samples environment actions and passes them to the simulator through the control functions \textsc{Initialize}, \textsc{Step}, and \textsc{IsTerminal}. The simulator applies these actions, updates its internal state, and outputs an indication whether an event in $E$ occurred and the likelihood of the latest state transition. The reward function transforms the simulator outputs into a reward to be passed back to the solver. The solver completes the loop by using the reward to choose the next action.

\section{AUTONOMOUS VEHICLES APPLICATION}\label{sec:astav}
This section defines the set of scenarios and metrics that we will use to evaluate the performance of the methods as well as the reward function used by the solvers.

\subsection{Simulator Design}\label{sec:algmet}
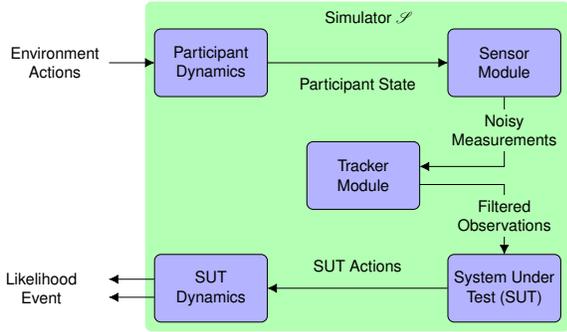
\begin{figure}[htbp]
	\centering
    \scalebox{0.60}{\input{SimStruct.tex}}
    \caption{The modular implementation of AST. The modules of the simulator can be easily swapped to test different scenarios, SUTs, or sensor configurations.}
	\label{fig:SIMStruct} 
\end{figure}
The driving algorithm, the sensors, the tracker, the solver, and the scenario definition are separate components in the framework. We use a modified version of the Intelligent Driver Model (IDM) as the SUT~\cite{PhysRevE621805}. If multiple IDM implementations were to be compared, it would be easy to swap them out and compare the results. The modularity gives AST the potential to be a useful benchmarking tool, or a batch testing method for autonomous systems.

The simulator architecture is shown in~\Cref{fig:SIMStruct}. The solver outputs the environment actions to the simulator, which is used to update non-SUT agents controlled by AST, called \emph{participants}. In our experiments, the only participants are pedestrians. The sensors receive the new participant states and output measurements augmented with the noise from the environment actions. The measurements are filtered by the tracker and passed to the SUT. The SUT, which is the driving model, decides how to maneuver the vehicle based on its observations. 

The SUT actions are used to update the state of the vehicle.  The simulator outputs the transition probability and event indicator to the reward function.  The current state of the simulator can be represented in different ways.  If the state of the simulator is fully observable, the simulator can provide it or an autoencoder processed version of it~\cite{kingma2013auto}. Otherwise, the history of previous actions can be used to represent the current state. The state representation, along with the reward of the previous step, are then input to the solver.

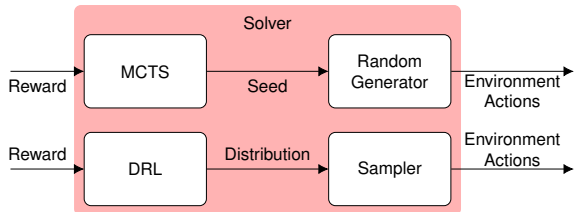
\begin{figure}[htbp]
	\centering
    \scalebox{0.65}{\input{SolverStruct.tex}}
    \caption{A comparison of the solver methods. MCTS uses a seed to control a random number generator. DRL outputs a distribution, which is then sampled. Both of these methods produce the environment actions.}
	\label{fig:SolverStruct} 
\end{figure}

Solvers use different procedures to generate the environment actions as shown in~\Cref{fig:SolverStruct}. MCTS outputs pseudorandom seeds, which are used to seed random number generators.  The environment actions are then sampled using these random number generators. DRL outputs a mean and standard deviation that characterize a multivariate Gaussian distribution. Environment actions are sampled from the distribution.
\subsection{Problem Formulation}\label{sec:setup}
\begin{figure}[htbp]
	
	\centering
    \begin{subfigure}{\columnwidth}
    \vspace*{0.25cm}
    \centering
    \includegraphics[width=0.70\columnwidth]{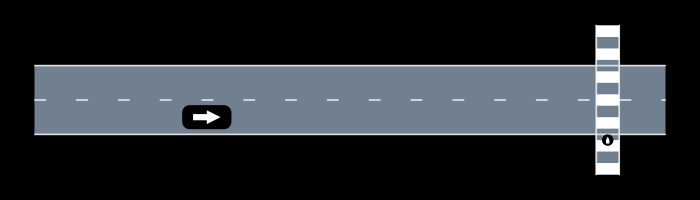}
    \caption{First scenario.}
	\label{fig:scenario1} 
    \end{subfigure}
    \begin{subfigure}{\columnwidth}
    \centering
    \includegraphics[width=0.75\columnwidth]{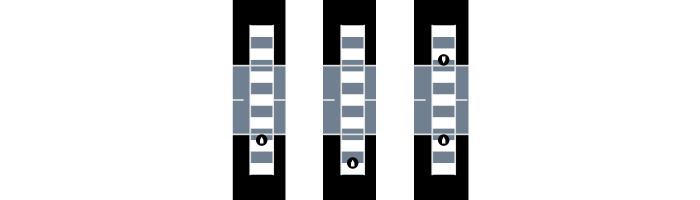}
    \caption{Initial pedestrian configurations for each scenario.}
	\label{fig:scenario2} 
    \end{subfigure}
    \caption{Three variations of initial configurations in the crosswalk scenario.}
    \label{fig:scenarios}
\end{figure}
To evaluate the effectiveness of AST as applied to autonomous vehicles, we stress test a vehicle in a set of scenarios at a pedestrian crosswalk, shown in \Cref{fig:scenarios}. The scenario is defined by a single autonomous vehicle approaching a crosswalk. The road has two lanes to model a regular neighborhood road, although there is no traffic in either direction for this specific example. We intentionally chose the simplest possible representative scenario for pedagogical reasons to illustrate the fundamental difference in scalability between DRL and MCTS. The road and crosswalk are sized according to California state regulations \cite{crosswalk}. The Cartesian origin is set at the intersection of the central vertical axis of the crosswalk and the central horizontal axis of the bottom lane, with the positive $x$ direction follows the direction of the arrow in \Cref{fig:scenarios}, and positive $y$ motion being towards the top side of the street. We test with different numbers of pedestrians, as well as with different starting states. The state of the $i$th pedestrian is $\vect{s}_{\text{ped}}^{( i )} = [v_x^{( i )},v_y^{( i )},x^{( i )},y^{( i )}]$ where 
\begin{itemize}
\item $v_x^{( i )},v_y^{( i )}$ are the $x$ and $y$ components of the velocity of the $i$th pedestrian.
\item $x^{( i )},y^{( i )}$ are the $x$ and $y$ components of the position of the $i$th pedestrian.
\end{itemize}
  We present data from each pedestrian from the three different variations of the scenario:
\begin{itemize}
\item 1 pedestrian, with initial state  $$\vect{s}_{\text{ped}}^{\left( 1 \right)} = \left[\SI{0.0}{\meter\per\second},\SI{1.4}{\meter\per\second},\SI{0.0}{\meter},\SI{-2.0}{\meter}\right]$$
\item 1 pedestrian, with initial state  $$\vect{s}_{\text{ped}}^{\left( 1 \right)} = \left[\SI{0.0}{\meter\per\second},\SI{1.4}{\meter\per\second},\SI{0.0}{\meter},\SI{-4.0}{\meter}\right]$$
\item 2 pedestrians, with initial state  $$\vect{s}_{\text{ped}}^{\left( 1 \right)} = \left[\SI{0.0}{\meter\per\second},\SI{1.4}{\meter\per\second},\SI{0.0}{\meter},\SI{-2.0}{\meter}\right]$$  $$\vect{s}_{\text{ped}}^{\left( 2 \right)} = \left[\SI{0.0}{\meter\per\second},\SI{-1.4}{\meter\per\second},\SI{0.0}{\meter},\SI{5.0}{\meter}\right]$$
\end{itemize}
shown in~\Cref{fig:scenario2}. The first scenario was chosen as a basic example to demonstrate AST. The second scenario was chosen to show that a different initial condition leads to different collision trajectories found. The third scenario shows the scalability of AST by including more participants in the scenario.
\subsubsection{Environment Models}
Both solvers use the same representation for environment actions. The environment action vector at each time step is $\vect{a}_{\text{env}} =  [\vect a^{(1)},\vect a^{(2)},\ldots,\vect a^{(n)}]$, where $n$ is the number of pedestrians.  For the $i$th pedestrian, $\vect{a}^{( i )} = [a_x^{( i )}, a_y^{( i )}, \epsilon_{v_x}^{( i )},\epsilon_{v_y}^{( i )},\epsilon_{x}^{( i )},\epsilon_{y}^{( i )} ]$ where 
\begin{itemize}
\item $a_x^{( i )},a_y^{( i )}$ are the $x$ and $y$ components of the $i$th pedestrian's acceleration.
\item $\epsilon_{v_x}^{( i )},\epsilon_{v_y}^{( i )}$ are the noise injected into the SUT measurement of the components of the $i$th pedestrian's velocity $v_x^{( i )}$ and $v_y^{( i )}$.
\item $\epsilon_{x}^{( i )},\epsilon_{y}^{( i )}$ are the noise injected into the SUT measurement of $x$ and $y$ components of the $i$th pedestrian's position.
\end{itemize}
AST controls both the pedestrian motion and the sensor noise, allowing it to search over both pedestrian actions and hardware failures to find the most likely collision. 

The inputs to the solvers vary slightly.  MCTS does not make use of the simulator's internal state, treating it entirely as a black box.  Instead, the AST implementation of MCTS differentiates states using a history of previous pseudorandom seeds~\cite{lee2015adaptive}. In contrast, DRL takes the simulation state as input. The simulation state is $\vect{s}_{\text{sim}} = [s_{\text{sim}}^{(1)},s_{\text{sim}}^{(2)},\ldots,s_{\text{sim}}^{(n)}]$. For the $i$th pedestrian,  $\vect{s}_{\text{sim}}^{( i )} =[\hat v_x^{( i )},\hat v_y^{( i )},\hat x^{( i )},\hat y^{( i )}]$ where
\begin{itemize}
\item $\hat v_x^{( i )},\hat v_y^{( i )}$ are the $x$ and $y$ components of the relative velocity between the SUT and the $i$th pedestrian.
\item $\hat x^{( i )},\hat y^{( i )}$ are the $x$ and $y$ components of the relative position between the SUT and the $i$th pedestrian.
\end{itemize}
At each time step, the pedestrian samples $\vect{a}^{(i)}$ (the procedure of this sampling differs slightly between solvers, but the representation of the action vector $\vect{a}^{(i)}$ is the same). To find the likelihood of $\vect{a}^{(i)}$, a model of the expected pedestrian action vector is needed. The model of the pedestrian is a multivariate Gaussian distribution $\mathcal{N}(\vect{\mu}_a, \vect\Sigma)$ where $\vect{\mu}_a$ is a zero-vector, and $\vect\Sigma$ is diagonal. Our pedestrian model is parameterized by $\sigma_{\text{aLat}}$, $\sigma_{\text{aLon}}$,  and $\sigma_{\text{noise}}$, which are the diagonal elements of the covariance matrix and correspond to lateral acceleration, longitudinal  acceleration, and sensor noise respectively. The values we use are: $\sigma_{\text{aLat}} = 0.01$, $\sigma_{\text{aLon}} = 0.1$, and $\sigma_{\text{noise}} = 0.1$. The acceleration parameters are designed to encourage the pedestrians to move across the street with some lateral movement. The assumption of the mean action being the zero-vector implies that, on average, a pedestrian holds course. In reality, this distribution could depend on the location of the pedestrian, where the vehicle is, the attitude or attention of the pedestrian, or other factors. Applying a more realistic pedestrian model is an avenue for future work.  The initial speed of the pedestrian is set to \SI{1.5}{\meter \per \second}, which is the average human walking speed.
\subsubsection{Sensor and Tracker Models}
The sensors of the SUT receive a vector of the participant state and output a vector of noisy measurements $\vect m = [m^{(1)},m^{(2)},\ldots,m^{(n)}]$. For the $i$th pedestrian,  $\vect m^{( i )} = \vect s_{ped}^{( i )} + \vect\epsilon^{(i)}$ where $ \vect\epsilon^{(i)} = [ \epsilon_{v_x}^{( i )},\epsilon_{v_y}^{( i )},\epsilon_{x}^{( i )},\epsilon_{y}^{( i )} ]$. The measurements are passed to an alpha-beta tracker \cite{alphabeta}, parameterized by  $\alpha_{\text{tracker}}$ and $\beta_{\text{tracker}}$, which returns the filtered versions of the measurements as the SUT's observations. We use the values $\alpha_{\text{tracker}} = 0.85$ and $\beta_{\text{tracker}} = 0.005$.
\subsubsection{System Under Test Model}
The SUT is based on the Intelligent Driver Model \cite{PhysRevE621805}. The IDM is designed to stay in one lane and safely follow traffic. To follow the rules around crosswalks, we set the desired velocity at $25$ miles per hour (\SI{11.17}{\meter \per \second}). If no lead vehicle is available to follow, the model maintains a desired velocity. We adapted the IDM for interacting with pedestrians by modifying it to treat the closest pedestrian in the road as the target vehicle. The IDM then tries to follow a safe distance behind the pedestrian based on their relative velocity, which results in the vehicle stopping at the crosswalk since the pedestrian's velocity along the $x$-axis is negligible. Our modified IDM is not a safe model; as we will show, ignoring any pedestrian outside of the road makes the vehicle vulnerable to being blindsided by people moving quickly from the curb. The goal of this paper, however, is to show that AST can effectively induce poor behavior in an autonomous driving algorithm, not to develop a safe algorithm ourselves. 
The SUT model receives a series of filtered observations  $\vect o = [o^{(1)},o^{(2)},\ldots,o^{(n)}]$. If there are pedestrians in the road, the SUT model uses the closest pedestrian to find $\vect s_{\text{SUT}} = [v_{\text{oth}}, s_{\text{headway}}]$ where
\begin{itemize}
\item $v_{\text{oth}}$ is the relative $x$ velocity between the SUT and the closest pedestrian.
\item $s_{\text{headway}}$ is the relative $x$ distance between the SUT and the closest pedestrian.
\end{itemize}
These factors determine the acceleration of the SUT in the next time step. 

\subsubsection{Modified Reward Function}
As a proxy for the probability of an action, we use the Mahalanobis distance \cite{mahalanobis1936generalised}, which is a measure of distance from the mean generalized for multivariate continuous distributions. The penalty for failing to find a collision is not actually $-\infty$, but instead a very large negative number. In addition, the penalty at the end of a no-collision case includes a component that is scaled by the distance ($\textsc{dist}$) between the pedestrian and the vehicle. The penalty encourages the pedestrian to end early trials closer to the vehicle, and leads to faster convergence. The reward function is modified from the previous version of AST \cite{lee2015adaptive} as follows:
\begin{equation}
R\left(s\right) = \left\{
        \begin{array}{ll}
            0 &  s \in E \\
            -10000 - 1000\times\textsc{dist}\left(\vect p_v,\vect p_p\right) &  s \notin E, t\geq T \\
            - \log \left(1 + M\left(a, \mu_a\mid s\right)\right)   &  s \notin E, t < T
        \end{array}
    \right.
\end{equation}
where $ M(a, \mu_a\mid s)$ is the Mahalanobis distance between the action $a$ and the expected action $\mu_a$ given the current state $s$. The distance between the vehicle position $\vect p_v$ and the closest pedestrian position $\vect p_p$ is given by the function $\textsc{dist}(\vect p_v,\vect p_p)$.
\subsubsection{Metrics}
We use two metrics to evaluate the AST algorithms. The first is the likelihood of the final collision trajectory output by the system.  The second metric is the number of calls to the step function. The goal of the second metric is to compare the efficiency of the two AST solvers. The separate implementations render both wall clock time and iterations inappropriate. The simulator update function (\textsc{Step}), which was the computational bottleneck, is used instead. This metric is agnostic to the implementation hardware, the algorithm used, and to the run-time of updating the simulation. 
\subsubsection{Solvers}
For MCTS, the parameters that control how much of the state space is explored are the depth, the horizon $T$, and the number of iterations. The depth and horizon are chosen to be equal so that the search and rollout stages explore the same scenario. We experimented with different values for the horizon $(50,75,100)$, and found that 100 was the minimum horizon that is sufficiently long to cover the scenario of interest. We used 2000 iterations. For additional detail on MCTS and DPW, see the paper by Lee et al. ~\cite{lee2015adaptive}.

For DRL, the results shown are obtained using a batch size of 4000. Experimentation showed that reducing the batch size any further resulted in too much variance during the trials. We use a step size of 0.1, and a discount factor of 0.99. The DRL approach is implemented using RLLAB~\cite{duan2016benchmarking}.
%

\section{RESULTS}\label{sec:res}

\begin{table*}[htbp]
\vspace*{0.25cm}
\ra{1.3}
\centering
\caption{Numerical results from both solvers. Reward without noise shows the reward of the MCTS path if sensor noise was set to zero, to illustrate the difficulty MCTS has with eliminating noise. DRL is able to find a more probable path than MCTS with a large reduction in calls to the \textsc{Step} function.}
\label{table:2}
\begin{tabularx}{\textwidth}{@{}XXXXcXX@{}}
\toprule
    & \multicolumn{3}{c}{MCTS} &  & \multicolumn{2}{c}{DRL} \\
\cmidrule{2-4} \cmidrule{6-7}
Scenario  & Calls to \textsc{Step} & Reward & Reward w/o noise && Calls to \textsc{Step} & Reward \\ 
\midrule
1 & \num{4.91e+08} & \num{-131} &\num{-71} && \num{8e+05} &\num{ -62} \\
2 & \num{1.85e+06} &\num{ -38} & \num{-15} &&  \num{8e+05} & \num{-1.7}  \\
3 & \num{1.61e+09} &\num{ -161} & \num{-104} && \num{1e+06} & \num{ -52}\\ 
\bottomrule
\end{tabularx}
\end{table*}
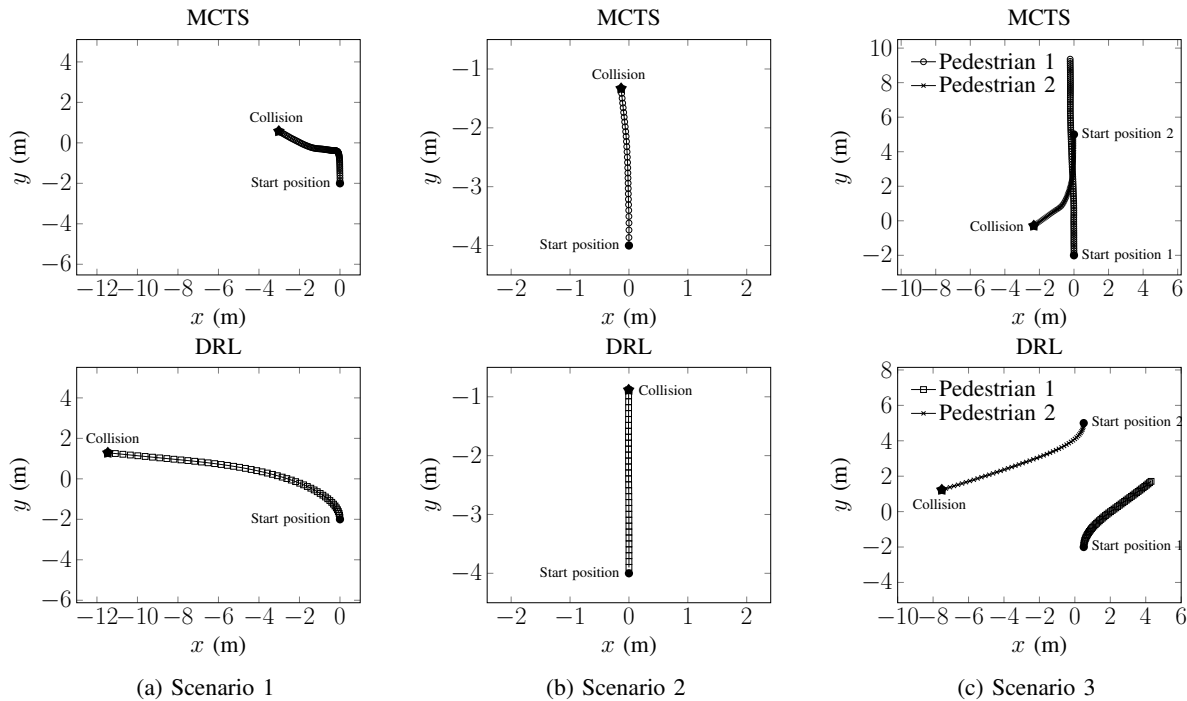
\begin{figure*}[h!]
	\centering
    \begin{subfigure}[t]{0.3\textwidth}
    	\scalebox{0.55}{\input{scen11_traj.tex}}
    \end{subfigure}
    \begin{subfigure}[t]{0.3\textwidth}
    	\scalebox{0.55}{\input{scen12_traj.tex}}
    \end{subfigure}
    \begin{subfigure}[t]{0.3\textwidth}
    	\scalebox{0.55}{\input{scen13_traj.tex}}
    \end{subfigure}\\
        \begin{subfigure}[t]{0.3\textwidth}
    	\scalebox{0.55}{\input{scen21_traj.tex}}
    	\caption{Scenario 1}
    \end{subfigure}
    \begin{subfigure}[t]{0.3\textwidth}
    	\scalebox{0.55}{\input{scen22_traj.tex}}
    	\caption{Scenario 2}
    \end{subfigure}
    \begin{subfigure}[t]{0.3\textwidth}
    	\scalebox{0.55}{\input{scen23_traj.tex}}
    	\caption{Scenario 3}
    \end{subfigure}
    \caption{Pedestrian motion trajectories for each scenario and algorithm. The collision point is the point of contact between the vehicle and the pedestrian. In scenario 3, pedestrian 1 does not collide with the vehicle.}
	\label{fig:Traj} 
\end{figure*}
The results show that both solvers are able to identify failure trajectories in an autonomous vehicle scenario. MCTS and DRL produce several situations where the vehicle collides with the pedestrian.  \Cref{table:2} shows the results for the three different scenarios. Both methods successfully converge to a solution in a tractable amount of simulator steps. AST is able to take advantage of the modified IDM's decision to ignore any pedestrian that is not in the road to find collisions. Although the likelihoods seem to vary greatly, much of this difference is due to MCTS having non-zero noise that adds up over the long horizon. The likelihood of pedestrian motion dominates that of sensor noise. As such, the noise should be very sparse, and the DRL solution reflects this. MCTS, however, has difficulty driving the noise to true zero. Instead, there are very small numbers in the noise vector throughout that can accumulate to a large probability error. We recomputed the reward as if the noise was 0 as a reference, which is also shown in \Cref{table:2}.
\subsection{Performance}
The number of calls to \textsc{Step} for MCTS is the number of calls required to find a collision in the rollouts. The algorithm itself does not commit to the actions that cause the collision until later. The number is provided to highlight the computational ability of the algorithm to find a collision, not the termination time. In addition, the number of calls presented for MCTS is the average over 100 single runs multiplied by 100 to represent the number of calls needed to have confidence in the results.
Across all scenarios, DRL consistently requires orders of magnitude fewer calls to \textsc{Step} than does MCTS. DRL converges to solutions with less than 1\% of the number of calls to \textsc{Step} required by MCTS despite the state and action spaces being very small. Theoretically, the scalability advantages of DRL should be even more apparent in a higher-dimensional problem. This advantage is supported by the fact that MCTS performs worse on the dual pedestrian scenario relative to both single pedestrian scenarios.
\subsection{Trajectories}

\Cref{fig:Traj} shows the pedestrian paths until the collision from each solver for scenarios 1, 2, and 3, respectively. The vehicle collides with the  pedestrian at the collision point. In scenario 1, both solvers send a single pedestrian into the road, and have the pedestrian move towards the vehicle to create a collision. However, the turn towards the vehicle is much more pronounced in MCTS, where the pedestrian comes to a near stop, before angling hard left and into the vehicle. DRL instead settles on a smoother path. The DRL path is slightly more likely because there is less acceleration needed. 
Scenario 2 has similar paths from the solvers. The pedestrians start at a point from which their mean action should create a collision. Both solvers identify this path quickly and the pedestrians take very little action. 

Both scenarios are relevant to scenario 3, which presents the largest difference. Because pedestrian 2 starts farther away from the vehicle, it has the more likely path to being hit by the vehicle as in scenario 2. In both solvers, the second pedestrian takes actions similar to scenario 1. However, there is a large difference in the first  pedestrian. In MCTS, the pedestrian holds course for a bit before aggressively accelerating towards the other side of the road. In DRL, the first pedestrian takes a slight turn to the right, and then holds a constant path from there. MCTS has less ability to minimize the effect of the first pedestrian on the total reward since using a single seed results in coupling the actions of the pedestrians. Hence, the first pedestrian has a different and less optimal trajectory than its counterpart in DRL. In DRL, the first pedestrian had a change of direction at first, causing the second pedestrian to be closer to the vehicle. Then the first pedestrian maintained a course with very little acceleration, minimizing the pedestrian's effect on the reward.

In scenarios 1 and 3, the blame of the collision is on the pedestrian, which does not inform any modifications to the SUT. However, in scenario 2, the blame is on the vehicle, since it does not check for pedestrians approaching the crosswalk until the pedestrians are in the crosswalk, which gives very short response time for the vehicle. The suggestion for avoiding a collision like scenario 2 is to expand the sensing range of the IDM to go beyond the curb of the road. The reason AST returns situations where the blame is not on the vehicle is that we define the subset of state space that we are interested in $E$ to be any collision. The kind of collisions reported by the examples shown in scenario 1 and 3 do not give the designer of the SUT any insight on how the SUT should be improved. The solution is to redefine the space of events of interest $E$ to be the subset of collisions where the responsibility of the collision was on the SUT. The definition of $E$ requires formal models of responsibility and blame in various road situations~\cite{mobileye}. Incorporating these models in the AST framework is an area of future work.
\section{CONCLUSIONS}\label{sec:conc}
This paper extended the adaptive stress testing methodology used before to test an aircraft collision avoidance system to autonomous vehicles. In addition, this paper introduced how to use deep reinforcement learning to improve the efficiency of adaptive stress testing. Deep reinforcement learning can find more-likely failure scenarios than Monte Carlo tree search, and it finds them more efficiently. Another contribution of this paper is a testing framework for autonomous vehicles that has modular components. By adapting one of the approaches, any manufacturer who wishes to test the sensor or decision system of a vehicle in simulation can use this framework. Further work will involve incorporating more realistic sensor and pedestrian models and imposing a tighter constraint on the definition of the events of interest.

\section*{Acknowledgements}
The authors would like to express their gratitude towards Tim Wheeler for his helpful comments and his assistance in using the Automotive Driving Models simulator, which this work is built on.
\bibliographystyle{IEEEtran}
\bibliography{egbib}

\end{document}

%% file: ASTStruct.tex
\begin{tikzpicture}[node distance=1.5cm,
    every node/.style={fill=white, font=\sffamily, text centered}, align=center]
	\node (sim)             [simulator]              {Simulator $\mathscr{S}$};
    \node (solver)          [solver, above of = sim, xshift = 0.8cm]              {Solver};
    \node (reward)          [reward, above of = sim, xshift = 5cm]              {Reward\\Function};  
  \draw[->]					(solver.west) -| node[text width=1cm, xshift = 0mm, yshift = 5mm, text centered, align=center] {Environment\\Actions} ($ (sim.north) - (15mm, 0) $);
  \draw[->] 	($(sim.east) + (0mm,2mm)$) -| node[text width=1.6cm, xshift = -17mm, yshift = 4mm] {Likelihood} ($ (reward.south) + (-2mm, 0mm) $);
  \draw[->] 	($(sim.east) + (0mm,-2mm)$) -| node[text width=1cm, xshift = -20mm, yshift = -4mm] {Event} ($ (reward.south) + (2mm, 0mm) $);
  \draw[->]		(reward.west) -- ++(0mm,0) -- node[text width=1cm, xshift = 0mm, yshift = 3mm] {Reward}(solver.east);
\end{tikzpicture}

%% file: SimStruct.tex
\begin{tikzpicture}[node distance=1.5cm,
    every node/.style={font=\sffamily}, align=center]
	\node (dyn1)			[module, xshift = 0cm, yshift = 0cm]			{Participant\\Dynamics};
    \node (sensor)			[module, right of = dyn1, xshift = 5cm, yshift = 0cm]			{Sensor\\Module};
    \node (tracker)				[module, below of = sensor, xshift = -3.125cm, yshift = -1cm]			{Tracker\\Module};
    \node (SUT)				[module, below of = tracker, xshift = 3.125cm, yshift = -1cm]			{System Under\\Test (SUT)};
    \node (dyn2)			[module, left of = SUT, xshift = -5cm, yshift = 0cm]			{SUT\\Dynamics};
    \path (dyn1) |- ++(3.5cm,1cm) node {Simulator $\mathscr{S}$};
    \node (spacer)			[module, fill=white, minimum size = 0.02cm, circle, above of = dyn1, inner sep = 0pt, outer sep= 0pt, yshift = 5mm] {};
    
	\draw[->]			($ (dyn1.west) + (-10mm,0mm)$)node [xshift = -12mm, fill=white]{Environment\\Actions} -- (dyn1.west);
    \draw[->]			(dyn1) -- node[yshift = -5mm]{Participant State} (sensor);
    \draw[->]			(sensor.south) |- node[fill = green!30, xshift = 0cm, yshift=8mm]{Noisy\\Measurements} ($ (tracker.east) + (0mm, 2mm) $);
    \draw[->]			($ (tracker.east) + (0mm, -2mm) $)-| node[fill = green!30, yshift=-7mm]{Filtered\\Observations} (SUT.north);
    \draw[->]			(SUT) -- node[yshift = 5mm]{SUT Actions} (dyn2);
    \draw[->]			($(dyn2.west) + (0mm,2mm)$) -- ++(-10mm,0cm) node[xshift = -15mm]{Likelihood};
    \draw[->]			($(dyn2.west) + (0,-2mm)$) -- ++(-10mm,0cm) node[xshift = -15mm]{Event};

    \begin{pgfonlayer}{background}
    	
    	\path[fill = green!30, rounded corners] ($(dyn1.north west) + (-2mm, 6mm)$) rectangle ($(SUT.south east) + (2mm, -2mm)$);

    \end{pgfonlayer}

  \end{tikzpicture}

%% file: SolverStruct.tex
\begin{tikzpicture}[node distance=1.5cm,
    every node/.style={font=\sffamily}, align=center]
	\node (mcts)		 [module2, xshift=-\xsplit cm, yshift = \ysplit cm]		{MCTS};
    \node (rand)		 [module2, xshift=\xsplit cm, yshift = \ysplit cm]		{Random\\Generator};
    \node (drl)		 [module2, xshift=-\xsplit cm, yshift = -\ysplit cm]		{DRL};
    \node (sampler)		 [module2, xshift=\xsplit cm, yshift = -\ysplit cm]		{Sampler};
    \node (name)		[yshift=2.0cm]			{Solver};
  	\draw[->] 		(mcts) -- node[yshift=-3mm]{Seed} (rand);
    \draw[->]		(drl) -- node[yshift=3mm]{Distribution} (sampler);
    \draw[->]		($ (mcts.west) + (-15mm,0) $) -- node[yshift=-3mm, fill=white, xshift=-2mm]{Reward} (mcts.west);
    \draw[->]		($ (drl.west) + (-15mm,0) $) -- node[yshift=3mm, fill=white, xshift=-2mm]{Reward} (drl.west);
    \draw[->] 		(rand.east) --node[yshift=-4mm]{Environment\\Actions} ++(2.5cm,0);
    \draw[->] 		(sampler.east) --node[yshift=4mm]{Environment\\Actions} ++(2.5cm,0);
  
    \begin{pgfonlayer}{background}
		\path[fill = red!30, rounded corners] ($(mcts.north west) + (-2mm, 6mm)$) rectangle ($(sampler.south east) + (2mm, -2mm)$);
    \end{pgfonlayer}
  \end{tikzpicture}

%% file: scen11_traj.tex
\begin{tikzpicture}[]
\begin{axis}
[ title = MCTS, axis equal = {true},xlabel={$x$ (m)},ylabel={$y$ (m)},xmin = {-13},xmax = {1}]\addplot+ [mark = o, fill = none,black] coordinates {
(0.0, -2.0)
(0.00035642999345136047, -1.8659298969195743)
(8.586768734429248e-5, -1.7372223383310739)
(-0.002058448320831586, -1.6125436829571502)
(-0.004864893477473036, -1.4959424528886724)
(-0.006795474496504399, -1.3866503333839386)
(-0.00822663954080921, -1.2828114743147558)
(-0.009885032698950622, -1.1827104941409858)
(-0.011590415628163979, -1.0884340187524417)
(-0.013137043051994979, -1.0019292594063698)
(-0.015173182542782514, -0.9222438023937674)
(-0.019247918838567597, -0.8497214945042435)
(-0.024706535733375366, -0.7852078111883953)
(-0.030197569030372386, -0.7268997515844937)
(-0.03654186698935264, -0.6742813936035255)
(-0.04444521245655549, -0.626612962956159)
(-0.05291213239536262, -0.5845469938815331)
(-0.0616038577668758, -0.5489737548498614)
(-0.07075489143893536, -0.5183365303012155)
(-0.07977498623913702, -0.4926547942826387)
(-0.08894575920155501, -0.47264625031212404)
(-0.09934739578390296, -0.4562850389977102)
(-0.11114623447962475, -0.44053313170709707)
(-0.12425861841174403, -0.427037501863917)
(-0.13925723901265527, -0.4178914145474657)
(-0.15550006217463325, -0.40915211299486876)
(-0.1730204491654155, -0.39937874621735237)
(-0.19219518875653335, -0.39278728881202785)
(-0.21254710885098882, -0.3891121628694614)
(-0.23401007954106803, -0.38621325960652214)
(-0.25748449297158715, -0.38261863593331386)
(-0.2832336523637814, -0.3786355796525473)
(-0.3100531360793579, -0.37590281987553226)
(-0.33820091550511494, -0.3744672229398706)
(-0.3691258600503351, -0.3739723681634452)
(-0.4026506426877878, -0.37232650725762)
(-0.4378968575043416, -0.36875313898901035)
(-0.4751660709316018, -0.3645637739676521)
(-0.5154984298522154, -0.358578893798227)
(-0.5596753051851737, -0.35075492546492754)
(-0.6058659540800899, -0.34429018636986214)
(-0.6533832445484717, -0.3382228228836079)
(-0.7024836537767907, -0.3319561804450387)
(-0.7527840666212934, -0.3263939613220658)
(-0.8047013576466295, -0.3178027981039192)
(-0.858356440041893, -0.3082345552358311)
(-0.9139009805375277, -0.303198424496938)
(-0.9699161328739225, -0.2992205883741156)
(-1.0260965531597908, -0.29383992410472215)
(-1.0837981211825811, -0.28890356207820733)
(-1.1433859551285055, -0.28394022947639624)
(-1.2041414734933955, -0.27609917629928643)
(-1.2664385103486027, -0.26418596292667945)
(-1.3312186291686023, -0.24926695685146322)
(-1.3980448600015944, -0.23173889499922834)
(-1.4667712884234947, -0.21073374394631622)
(-1.5376364212567926, -0.18454703619818538)
(-1.610517411908367, -0.15611798563943172)
(-1.6845185394988296, -0.12616900328262906)
(-1.759122692066025, -0.09326702876176962)
(-1.8343495271813968, -0.06031189749533339)
(-1.9108638896639736, -0.023492946796986425)
(-1.9886466495296964, 0.017850781848032925)
(-2.0667570955370986, 0.058407180991063595)
(-2.145276019169309, 0.09765493638479254)
(-2.2241776620728366, 0.13800095854061595)
(-2.3032934152135196, 0.17896339765341934)
(-2.38273915827858, 0.21761718982407605)
(-2.461584289811174, 0.25967354407524984)
(-2.5398058403655095, 0.3058786049846852)
(-2.6191857639161578, 0.35085384535168573)
(-2.6990297300815316, 0.3957890878535951)
(-2.778443539079056, 0.4416426304739146)
(-2.8584615472268573, 0.4861026774537919)
(-2.938872786751854, 0.5302705503403508)
(-3.0197100781732984, 0.5766802972550068)
};
\node[label={[left,yshift=0ex,xshift=0em]180:{\normalsize{Start position}}},circle,fill,inner sep=2pt] at (axis cs:0,-2) {};
\node[label={[right,yshift=2.2ex,xshift=-2em]180:{\normalsize{Collision}}},star,fill,inner sep=2pt] at (axis cs:-3.0197100781732984, 0.5766802972550068) {};
\end{axis}

\end{tikzpicture}

%% file: scen12_traj.tex
\begin{tikzpicture}[]
\begin{axis}
[title = MCTS, axis equal = {true},xlabel={$x$ (m)},ylabel={$y$ (m)},xmin={-2},xmax={2},ymin={-4.5}, ymax={-0.5}]\addplot+ [mark = o, fill = none,black] coordinates {
(0.0, -4.0)
(-1.5947378748584142e-6, -3.864370452310259)
(0.0006158389424357696, -3.7370188639445225)
(0.0007455830313880912, -3.617422908846752)
(-0.0006675277596587204, -3.5056238668025834)
(-0.003233500175189616, -3.4024882406421164)
(-0.006303883642038682, -3.3068918746411295)
(-0.008794722905526634, -3.2140178721104657)
(-0.010621189796949534, -3.1218851656806534)
(-0.011955147969488027, -3.030503894851879)
(-0.013688846424898513, -2.9404736022786007)
(-0.016018331960406237, -2.8532698298571177)
(-0.0180224870983309, -2.7671627644557795)
(-0.020332595711145072, -2.6790304488619507)
(-0.02344850426560036, -2.5890767936425414)
(-0.027347485824425005, -2.498705265412543)
(-0.03199714479143079, -2.4083837988013017)
(-0.03679036639588307, -2.319510709392059)
(-0.04172715063778184, -2.2320859971848144)
(-0.04781426972759785, -2.1461105753240375)
(-0.055147029654107275, -2.0637117502559343)
(-0.06335138721581952, -1.9819671099548226)
(-0.07208432875463618, -1.8991971536057424)
(-0.0807116830971109, -1.818643555104308)
(-0.08943110168442908, -1.7388241805643005)
(-0.09800924658320315, -1.659235006655059)
(-0.10647850760233472, -1.5793004817803649)
(-0.1155373464811706, -1.4976283946254498)
(-0.12437253058529209, -1.4144345002378018)
(-0.1328093348347523, -1.3302582592665892)
};
\node[label={[left,yshift=0ex,xshift=0em]180:{\normalsize{Start position}}},circle,fill,inner sep=2pt] at (axis cs:0,-4) {};
\node[label={[right,yshift=2.2ex,xshift=-2em]180:{\normalsize{Collision}}},star,fill,inner sep=2pt] at (axis cs:-0.1328093348347523, -1.3302582592665892) {};
\end{axis}

\end{tikzpicture}

%% file: scen13_traj.tex
\begin{tikzpicture}[]
\begin{axis}[title = MCTS, legend pos = {north west},axis equal = {true},xlabel={$x$ (m)},ylabel={$y$ (m)},legend style={draw=none},xmin = {-10}, xmax = {6}]\addplot+ [mark = o, fill = none,black] coordinates {
(0.0, -2.0)
(-0.0003058319552648728, -1.861614367667016)
(-0.0014208967139114586, -1.7248425539884342)
(-0.00337259618681525, -1.5905729367951218)
(-0.004826450650648078, -1.459712268988153)
(-0.006140393130266951, -1.332189647969193)
(-0.0073452584214573746, -1.2062426541650848)
(-0.007547363376776532, -1.0789690534381304)
(-0.007272530253578637, -0.9529071160213181)
(-0.007016851473301293, -0.8291517251567244)
(-0.007407124410919499, -0.7059225693848781)
(-0.008585616215396873, -0.5815986425744355)
(-0.010178484000335594, -0.454913379597766)
(-0.011634932059636033, -0.32414437438052124)
(-0.013072721853427232, -0.19107822189922796)
(-0.013955541793724848, -0.05896961192308625)
(-0.013666502417385361, 0.0721086612656201)
(-0.012303640781772604, 0.2017807981115034)
(-0.010795115911562651, 0.33222265762397596)
(-0.009911232497686733, 0.4667466397017592)
(-0.009630652308845813, 0.6018574960450676)
(-0.011150558821824625, 0.7378158879832002)
(-0.013960236970032128, 0.8747579988440934)
(-0.01630848239647343, 1.0115382457797093)
(-0.019944130365411913, 1.1498655313199055)
(-0.02496615911423552, 1.2882810445148163)
(-0.02902731514358251, 1.4272975207775573)
(-0.03315631684885521, 1.5686515614440681)
(-0.03810873943654737, 1.7095961063432767)
(-0.042341715397764, 1.8470906523378474)
(-0.04652512594242309, 1.98362293271322)
(-0.05205748719647986, 2.1216158328683647)
(-0.05838685083170906, 2.261778301784683)
(-0.0647664256890079, 2.4048940967689774)
(-0.07185409022864134, 2.5487217918444687)
(-0.0795412926496128, 2.691475363969589)
(-0.08712298684837114, 2.833545397182979)
(-0.09508306193931342, 2.9758434992463316)
(-0.10264102877854284, 3.121097080430993)
(-0.10967549438076642, 3.2699893538606455)
(-0.11694353224028059, 3.419766466779663)
(-0.12383901801748522, 3.5688861955276394)
(-0.1296927175017288, 3.7185098425840035)
(-0.13411334317585083, 3.868309446090792)
(-0.13815207573555122, 4.018311441498045)
(-0.14276502152269016, 4.166373373429099)
(-0.14752214511982475, 4.310026780627557)
(-0.1524832277071511, 4.450920968200791)
(-0.15809014485662107, 4.591395259440004)
(-0.16391956075023076, 4.730525749059351)
(-0.16930328681069293, 4.865872298718729)
(-0.17464477883464724, 4.998153605234216)
(-0.1800917248056268, 5.128870708029375)
(-0.1854269371641227, 5.259916422790223)
(-0.18977931715448373, 5.395590498527081)
(-0.1929633083749929, 5.534546201557509)
(-0.1959122371048254, 5.673745602314522)
(-0.19912387316008476, 5.812561254030545)
(-0.2023922697999795, 5.950960473960965)
(-0.2052200581835767, 6.088108464344861)
(-0.20767470954249234, 6.221885441112358)
(-0.2092261804001798, 6.3532524023700905)
(-0.20941750828827155, 6.484334252581576)
(-0.209126316271417, 6.61636461417466)
(-0.20922075072636165, 6.7487963838045175)
(-0.20964467351274996, 6.883080404068407)
(-0.20959960744261794, 7.0157034730735965)
(-0.20885366806122083, 7.145016909544811)
(-0.20799431927923773, 7.27221552455776)
(-0.20760355053106674, 7.39586137820673)
(-0.2076194707224198, 7.515641987639301)
(-0.20707855801875774, 7.633512259995456)
(-0.20622702492167122, 7.75170719594759)
(-0.20573743875851, 7.871104519651707)
(-0.20571240666858684, 7.99156250774848)
(-0.20662580640795314, 8.112119637739328)
(-0.20772493195653874, 8.233166121344157)
(-0.20867289847647716, 8.3558558369063)
(-0.20930390782372058, 8.480644439413481)
(-0.20983926750590598, 8.606796831238462)
(-0.21133723613019573, 8.734475385443304)
(-0.21348192905931754, 8.860864368221726)
(-0.21561488663468228, 8.986004227200493)
(-0.21757405434554725, 9.110095615517652)
(-0.21921491655212078, 9.23362709946664)
(-0.22087109161218196, 9.358477529482064)
};
\addlegendentry{Pedestrian 1}
\addplot+ [mark = x, fill = none,black]coordinates {
(0.0, 5.0)
(0.00013129211906961997, 4.863102408957329)
(0.001019388749773544, 4.733727861970034)
(0.0024845024031883533, 4.610786218448735)
(0.0026831271238555838, 4.492373584925093)
(0.0016870372917643206, 4.377867980004483)
(0.00026337524184905895, 4.269433222988939)
(-0.0018874942360586555, 4.16859307144348)
(-0.004251164006854459, 4.073175099963898)
(-0.007191507549425159, 3.9843012648499636)
(-0.010036754462936961, 3.903620631190762)
(-0.012143578844505946, 3.8292607174525575)
(-0.013618508880749247, 3.7571830868492118)
(-0.014870411226636123, 3.6873439842909033)
(-0.016168513536536623, 3.6226732802437187)
(-0.016981516161842854, 3.5624550825112005)
(-0.018925818577411747, 3.5037889721352817)
(-0.022604899950740182, 3.4442790965222017)
(-0.027107264367254122, 3.3845541366529934)
(-0.03145722660909036, 3.324641521112034)
(-0.034510866571826705, 3.26377597600632)
(-0.03711677874261075, 3.2026854609100264)
(-0.040434167093181886, 3.138700684446194)
(-0.04469779962751012, 3.0715634500799096)
(-0.0499777913154771, 3.0048082710174615)
(-0.055436039979124094, 2.9377806341081474)
(-0.06167334134608339, 2.8686907704022513)
(-0.06870742973310587, 2.800297281790769)
(-0.07660123233705782, 2.7342845016990163)
(-0.08561314517816747, 2.6696910494429638)
(-0.09443059701293194, 2.605480232610473)
(-0.10366120298097642, 2.542395545227257)
(-0.11388908320234951, 2.4800457932528133)
(-0.12502820514536048, 2.418536165409761)
(-0.13711616992727643, 2.3584427913749284)
(-0.15064451678370325, 2.298451392094856)
(-0.16652954558880104, 2.2370463897334565)
(-0.1837343585952292, 2.1740242394814313)
(-0.2010689693583377, 2.108895973319121)
(-0.21914377918786304, 2.0391444303811124)
(-0.23778106088225176, 1.9661426903552552)
(-0.2575818143732102, 1.8916337397590413)
(-0.2797066129118319, 1.816220511854004)
(-0.302814318559137, 1.741090572622312)
(-0.3259561810481183, 1.6674935271737885)
(-0.34964160299554153, 1.596438161123551)
(-0.3746620696828831, 1.5284219867254771)
(-0.40042403915437685, 1.4610098869842973)
(-0.42644058081089187, 1.392304986151502)
(-0.4534596973103578, 1.324316318958652)
(-0.48193010218638505, 1.259135858251958)
(-0.5115422098855222, 1.196438997192936)
(-0.5417375896754418, 1.1345392980435713)
(-0.5739224180194384, 1.0752655134685138)
(-0.6078496221663482, 1.019449130923877)
(-0.6427486172973794, 0.9667640471604876)
(-0.6785933918615895, 0.9171617781154088)
(-0.7148269691768504, 0.8714297219695887)
(-0.7521296347686944, 0.8312480290566834)
(-0.7913987946826218, 0.7961155962358158)
(-0.8332964244545746, 0.7637475514575885)
(-0.8772511018961272, 0.7305726523194852)
(-0.9219650803292155, 0.6976359869325207)
(-0.9679239780641995, 0.6682174944209424)
(-1.0161272546796443, 0.6390811226036069)
(-1.0660594052119392, 0.6055591885823058)
(-1.1174297374129896, 0.5694880183533929)
(-1.1703077337576757, 0.534591811845791)
(-1.2246928686585996, 0.4983632137234976)
(-1.2801263833882528, 0.4589143693620234)
(-1.3365884995632697, 0.41936640447941204)
(-1.3951853253352977, 0.3767080281585639)
(-1.4554694511771316, 0.3326773269364251)
(-1.516729585523133, 0.28885688914692736)
(-1.5788426219819585, 0.24334479030856546)
(-1.6420644900170722, 0.1974742387571455)
(-1.706646426968641, 0.1502769204414791)
(-1.7724175099583444, 0.10328738987747399)
(-1.839319155789183, 0.0554738345439445)
(-1.9074063381812956, 0.006086609053593828)
(-1.9762689153162631, -0.04182829775973638)
(-2.045501313867741, -0.0867117093184124)
(-2.1157932211356734, -0.13188249956914674)
(-2.186178164694324, -0.18118993543724765)
(-2.2560671731642534, -0.23282939689316579)
(-2.3258793526592387, -0.2849434606599403)
};
\node[label={[right,yshift=0ex,xshift=0.5em]180:{\normalsize{Start position 1}}},circle,fill,inner sep=2pt] at (axis cs:0,-2) {};
\node[label={[right,yshift=0ex,xshift=0.5em]180:{\normalsize{Start position 2}}},circle,fill,inner sep=2pt] at (axis cs:0,5) {};
\addlegendentry{Pedestrian 2}
\node[label={[left,yshift=0ex,xshift=0em]180:{\normalsize{Collision}}},star,fill,inner sep=2pt] at (axis cs:-2.3258793526592387, -0.2849434606599403) {};
\end{axis}

\end{tikzpicture}

%% file: scen21_traj.tex
\begin{tikzpicture}[]
\begin{axis}[ title = DRL, axis equal = {true},xlabel={$x$ (m)},ylabel={$y$ (m)},xmin = {-13},xmax = {1}]\addplot+[mark = square, fill = none,black] coordinates {
(-0.005000000000000001, -1.9007861739473404)
(-0.020000000000000004, -1.8038375449631432)
(-0.045000000000000005, -1.7094896595358984)
(-0.08000000000000002, -1.617027055246635)
(-0.12500000000000003, -1.5248246256645983)
(-0.18000000000000005, -1.4322052727668628)
(-0.24500000000000005, -1.3401825342912599)
(-0.32000000000000006, -1.2498957113165146)
(-0.4050000000000001, -1.161482579854066)
(-0.5000000000000001, -1.0735055373489848)
(-0.6050000000000001, -0.9848276751413338)
(-0.7200000000000001, -0.8959082035312004)
(-0.8444091112965983, -0.8081176155870756)
(-0.9782128454200019, -0.7212173567191122)
(-1.1220020910736124, -0.6343982735186415)
(-1.275791336727223, -0.5478197036968262)
(-1.4392505842821253, -0.4623860351028352)
(-1.6123798337383197, -0.379032878631945)
(-1.7955090831945142, -0.2969527942487495)
(-1.9886383326507087, -0.21606408568710672)
(-2.19127157229644, -0.13732251541669027)
(-2.403408802131708, -0.061018555620627524)
(-2.624544108011713, 0.012820427405627582)
(-2.8537671366779063, 0.0849056334196109)
(-3.0916284270696304, 0.1545611634141908)
(-3.335781052957949, 0.22119474008363363)
(-3.586625129153283, 0.28526223386188515)
(-3.845758414278429, 0.34670992445454457)
(-4.11160370045121, 0.40583208889355593)
(-4.385689847616567, 0.4628026932271675)
(-4.66852538726317, 0.5174364287700317)
(-4.958977204045072, 0.5697009366561536)
(-5.256451194901603, 0.6190890840437395)
(-5.560339845562125, 0.6654217443556607)
(-5.870306578050019, 0.7093514748648267)
(-6.186832359927392, 0.751914312789003)
(-6.510929435379563, 0.7925800971032977)
(-6.840133141039416, 0.8308312207316478)
(-7.172633882292547, 0.8673344840736008)
(-7.510642354306377, 0.902200092522494)
(-7.8528059204560074, 0.9356462106551351)
(-8.19811668299282, 0.9684309174482068)
(-8.547313770563388, 1.0018313955157556)
(-8.900308431134885, 1.0359254885832392)
(-9.257759263655162, 1.0706071118791167)
(-9.618961549372246, 1.1057310093618142)
(-9.98286815314072, 1.140827825039901)
(-10.350074588029214, 1.1768259116858366)
(-10.719762275960925, 1.2137967255870685)
(-11.090671163703204, 1.2506135256116193)
(-11.463094906786496, 1.2866249015604592)
};
\node[label={[left,yshift=0ex,xshift=0em]180:{\normalsize{Start position}}},circle,fill,inner sep=2pt] at (axis cs:0,-2) {};
\node[label={[right,yshift=2.2ex,xshift=-1.5em]180:{\normalsize{Collision}}},star,fill,inner sep=2pt] at (axis cs:-11.463094906786496, 1.2866249015604592) {};
\end{axis}

\end{tikzpicture}

%% file: scen22_traj.tex
\begin{tikzpicture}[]
\begin{axis}[title = DRL,axis equal = {true},xlabel={$x$ (m)},ylabel={$y$ (m)},xmin={-2},xmax={2},ymin={-4.5}, ymax={-0.5}]\addplot+[mark = square, fill = none,black] coordinates {
(0.0001851894013765132, -3.8999873288770948)
(0.00034349248939968054, -3.7999134808984216)
(0.00032706604487303236, -3.699745537626826)
(0.00042643741864109813, -3.59951578390042)
(0.0005146109764544412, -3.4992288345418796)
(0.0002642456230446879, -3.398874850598305)
(-8.947836494284405e-5, -3.298529522497733)
(-0.00037213747328768663, -3.198281018977833)
(-0.0007394050243492946, -3.0980249764527725)
(-0.0010876884379385256, -2.997711823420998)
(-0.001431744743750889, -2.8973112343774634)
(-0.0018732092489167434, -2.796839329299194)
(-0.002536743324895169, -2.696251403590259)
(-0.003495002040828585, -2.5955120269441925)
(-0.004629001914415041, -2.4948382793909936)
(-0.005554472634822894, -2.394232179273919)
(-0.006024545042229463, -2.2935355459944873)
(-0.00626543489102782, -2.1926835875533555)
(-0.006549440862980432, -2.091703888804769)
(-0.0067779359970993385, -1.9905955745649233)
(-0.006937389826858223, -1.889477145132795)
(-0.007117740526619195, -1.7884035476433107)
(-0.007312157819831814, -1.687384162930995)
(-0.007316198706203947, -1.5865554316258805)
(-0.007004157500140324, -1.4859084537818328)
(-0.006516379987893413, -1.3853366161391882)
(-0.005972727828593515, -1.2847146714600832)
(-0.005659479132701902, -1.1840661805426065)
(-0.00545992697269649, -1.0834615651978265)
(-0.00497995252486461, -0.9828811396631466)
(-0.004650728127522779, -0.8822879205224583)
};
\node[label={[left,yshift=0ex,xshift=0em]180:{\normalsize{Start position}}},circle,fill,inner sep=2pt] at (axis cs:0,-4) {};
\node[label={[right,yshift=0ex,xshift=0.7em]180:{\normalsize{Collision}}},star,fill,inner sep=2pt] at (axis cs:-0.004650728127522779, -0.8822879205224583) {};
\end{axis}

\end{tikzpicture}

%% file: scen23_traj.tex
\begin{tikzpicture}[]
\begin{axis}[title = DRL, legend pos = {north west},axis equal = {true},xlabel={$x$ (m)},ylabel={$y$ (m)},legend style={draw=none},xmin = {-10}, xmax = {6}]\addplot+[mark = square, fill = none,black] coordinates {
(0.5043555665682921, -1.901071365479615)
(0.5160802803115383, -1.8051158917092787)
(0.5340983066000827, -1.7121737229940277)
(0.5591513612973876, -1.6215813802532877)
(0.5909834855191456, -1.533251264546035)
(0.629064833724207, -1.4470174330087602)
(0.6718598620836729, -1.3624497004655518)
(0.7186285767683758, -1.2794164870595508)
(0.7686697747933042, -1.1983855722454482)
(0.8214411830958095, -1.1191927659736711)
(0.8776834159086305, -1.041172845454422)
(0.9376244737286504, -0.963792899141561)
(1.002578206420216, -0.8873649029800829)
(1.0733825453276906, -0.8115558476453724)
(1.1482318839758783, -0.7360482330936696)
(1.2244047425689437, -0.6617401760146304)
(1.3025802491632303, -0.588204951244492)
(1.3829697002214087, -0.5143502830280013)
(1.465336142117906, -0.4406577875953419)
(1.5500060863620562, -0.3676931148991495)
(1.6358577079953205, -0.29504243800219654)
(1.7236621120529496, -0.2220940012126273)
(1.8156530258606498, -0.14887068532881773)
(1.9115736421920277, -0.07595871207670465)
(2.009420066041569, -0.0034730144334266244)
(2.108736456009557, 0.06867932445009771)
(2.208853648648792, 0.14076006003423078)
(2.3098630966406812, 0.21312808505092767)
(2.4097605614111375, 0.28579262679800976)
(2.50849626014938, 0.3585069645441614)
(2.607601088757171, 0.43097741508502685)
(2.7064649644937284, 0.5030249801173513)
(2.8040362925781745, 0.5748339946688414)
(2.900740582886622, 0.6464131240756364)
(2.9990382654815337, 0.717524990228343)
(3.0971647000492792, 0.7880520457216192)
(3.193634582246332, 0.8583541068903734)
(3.289976329455642, 0.9287632213672347)
(3.3851349356150426, 0.9993359064737748)
(3.4785817813604836, 1.0702705894608417)
(3.5732826757689264, 1.1410241965122616)
(3.668259399061584, 1.2114279939769523)
(3.762412037453263, 1.2817969820285222)
(3.8551721182854375, 1.3522421933270383)
(3.945951491128912, 1.4228575230313483)
(4.035751175401665, 1.493567773266588)
(4.12450688843971, 1.5647644306405102)
(4.212144281328071, 1.6363371855384192)
(4.299430733155054, 1.7080345678728968)
};
\addlegendentry{Pedestrian 1}
\addplot+[mark = x, fill = none,black] coordinates {
(0.495, 4.901412796998882)
(0.48, 4.805731642772224)
(0.45499999999999996, 4.712399954493189)
(0.41999999999999993, 4.620719799724278)
(0.37499999999999994, 4.531144713230788)
(0.31999999999999995, 4.443287321780646)
(0.25499999999999995, 4.356698776167835)
(0.17999999999999994, 4.271750219281299)
(0.09499999999999993, 4.188052086904039)
(-6.938893903907228e-17, 4.105443419633223)
(-0.10457053656109397, 4.023473963761074)
(-0.21823052280739969, 3.942239386617705)
(-0.3406376644530271, 3.8622041583187676)
(-0.47000694157952877, 3.783194921234998)
(-0.6062110733037502, 3.704817146463402)
(-0.7513369011916092, 3.626975490068364)
(-0.9049530195820021, 3.5501015334757646)
(-1.0644962331520358, 3.473909630506189)
(-1.2296625218177573, 3.398314180181978)
(-1.4006649094674162, 3.323418345403616)
(-1.5777822381627729, 3.2490483039550573)
(-1.7616044563660562, 3.175062264731213)
(-1.951419118589361, 3.1009120145947384)
(-2.1465032958192363, 3.0264773756800927)
(-2.3438019131659433, 2.952921751940418)
(-2.5431901328352633, 2.880181643262624)
(-2.745296599267826, 2.807236152916303)
(-2.9481398221651935, 2.734216561361504)
(-3.1519212722017413, 2.6615367260672524)
(-3.3566575554298836, 2.5891298691911353)
(-3.5630618815418305, 2.5171791389903846)
(-3.7704096636187545, 2.4456190494137067)
(-3.978143011764398, 2.374109452705327)
(-4.188305285929981, 2.302237682383855)
(-4.401779349644231, 2.229563740219031)
(-4.616906646364821, 2.15686599602717)
(-4.832777261846036, 2.0850589192631728)
(-5.050911372378249, 2.01385791628973)
(-5.269892489059955, 1.9425230636014084)
(-5.4869951616789985, 1.8710963603615003)
(-5.703383889491085, 1.8001110935396)
(-5.922239869818656, 1.7295494045497153)
(-6.143235160956024, 1.6585889434436194)
(-6.366604777137088, 1.5875403303284994)
(-6.592088485686633, 1.516932504144109)
(-6.818765276959025, 1.446198272704602)
(-7.046911255939527, 1.3749171139826815)
(-7.276801870236716, 1.3028948113899292)
(-7.508277548335256, 1.2309071207586124)
};
\node[label={[right,yshift=0ex,xshift=0.5em]180:{\normalsize{Start position 1}}},circle,fill,inner sep=2pt] at (axis cs:0.5,-2) {};
\node[label={[right,yshift=0ex,xshift=0.5em]180:{\normalsize{Start position 2}}},circle,fill,inner sep=2pt] at (axis cs:0.5,5) {};
\addlegendentry{Pedestrian 2}
\node[label={[right,yshift=-2.2ex,xshift=-2em]180:{\normalsize{Collision}}},star,fill,inner sep=2pt] at (axis cs:-7.508277548335256, 1.2309071207586124) {};
\end{axis}

\end{tikzpicture}